\documentclass[sigconf, screen]{acmart}
\usepackage{graphicx}
\usepackage{array} 
\usepackage{pifont}
\usepackage{makecell}
\usepackage{xcolor}
\usepackage{multirow}
\usepackage{cuted}
\usepackage{capt-of}
\usepackage{booktabs}
\AtBeginDocument{%
  }

\setcopyright{acmlicensed}
\copyrightyear{2018}
\acmYear{2018}
\acmDOI{XXXXXXX.XXXXXXX}
\acmConference[Conference acronym 'XX]{Make sure to enter the correct
  conference title from your rights confirmation email}{June 03--05,
  2018}{Woodstock, NY}
\acmISBN{978-1-4503-XXXX-X/2018/06}

\begin{document}

\title{Beyond Fidelity: Semantic Similarity Assessment in Low-Level Image Processing}

\author{Runjie Wang}
\orcid{0009-0008-9326-2582}
\affiliation{%
  \institution{Fuzhou University}
  \city{Fuzhou}
  \state{Fujian}
  \country{China}
  }
\email{wangrunjie2023@163.com}

\author{Weiling Chen}
\affiliation{%
  \institution{Fuzhou University}
  \city{Fuzhou}
  \state{Fujian}
  \country{China}
  }
\email{weiling.chen@fzu.edu.cn}

\author{Tiesong Zhao}
\authornote{Corresponding author: Tiesong Zhao}
\affiliation{%
  \institution{Fuzhou University}
  \city{Fuzhou}
  \state{Fujian}
  \country{China}
}
\email{t.zhao@fzu.edu.cn}

\author{Chang-Wen Chen}
\affiliation{%
  \institution{The Hong Kong Polytechnic University}
  \city{Hong Kong}
  \country{China}
}
\email{changwen.chen@polyu.edu.hk}

\renewcommand{\shortauthors}{Trovato et al.}


\begin{abstract}
Low-level image processing has long been evaluated mainly from the perspective of visual fidelity. However, with the rise of deep learning and generative models, processed images may preserve perceptual quality while altering semantic content, making conventional Image Quality Assessment (IQA) insufficient for semantic-level assessment. In this paper, we formalize \textit{Semantic Similarity} as a new evaluation task for low-level image processing, aimed at measuring whether semantic content is preserved after processing. We further present a structured formulation of image semantics based on semantic entities and their relations, and discuss the desired properties and constraints of a valid semantic similarity index. Based on this formulation, we propose Triplet-based Semantic Similarity Score (T3S), which models image semantics through foreground entities, background entities, and relations. T3S combines semantic entity extraction, foreground-background disentanglement, and open-world class/relation modeling. Experiments on COCO and SPA-Data show that T3S consistently outperforms existing fidelity-oriented metrics and representative semantic-level baselines, while better reflecting progressive semantic changes under diverse degradations. These results highlight the importance of semantic assessment in modern low-level vision.
\end{abstract}

\ccsdesc[300]{Computing methodologies~Computer vision}

\keywords{Semantic Similarity, Low-Level Image Processing, Image Quality Assessment, Semantic Preservation, Triplet-based Modeling}

\received{20 February 2007}
\received[revised]{12 March 2009}
\received[accepted]{5 June 2009}

\maketitle

\section{Introduction}

\begin{figure}[t] 
\centering 
\includegraphics[width=0.45\textwidth, keepaspectratio]{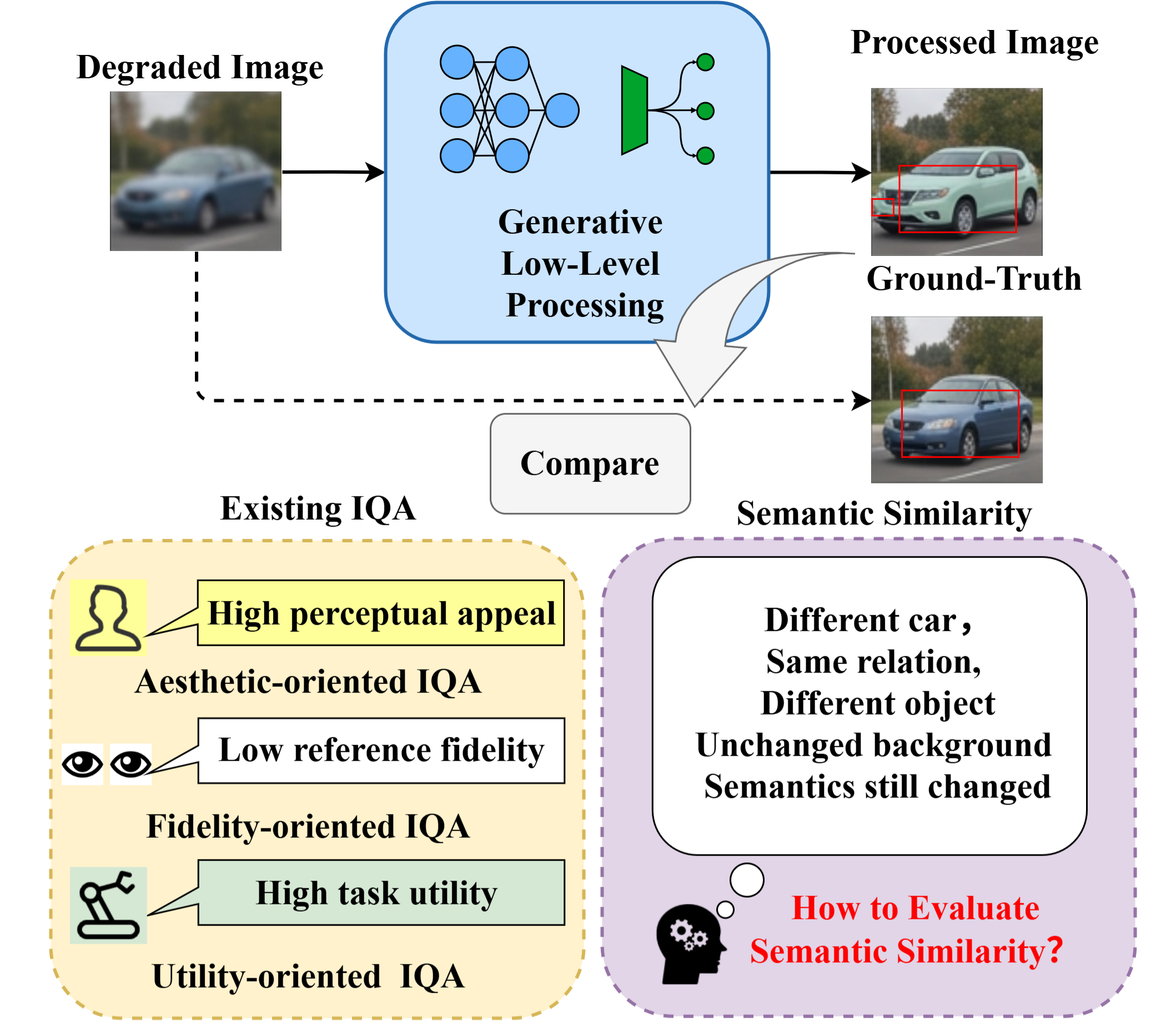} 
\caption{How to evaluate Semantic Similarity? Existing IQA methods cannot fully capture semantic preservation. In particular, generative restoration may introduce semantic distortions, which motivates Semantic similarity as a new task in low-level image processing.} 
\label{fig:pc} 
\end{figure}
Low-level image processing aims to restore a degraded image. Although these operations are usually performed at the pixel level, they often cause perceptual changes that can be noticed by human viewers. Therefore, it is necessary to measure these changes in order to evaluate the performance of image processing algorithms. In traditional IQA, different metrics are used for this purpose. Existing IQA methods can be broadly grouped into three categories \cite{Utility1}: Aesthetic-oriented IQA, Utility-oriented IQA, and Fidelity-oriented IQA. Aesthetic-oriented IQA focuses more on visual preference than on faithful content preservation. Utility-oriented IQA depends strongly on specific downstream tasks such as detection, recognition, and segmentation \cite{CVPR25iqaLiImage}. These two types of IQA are not ideal as general evaluation methods for low-level image processing. In contrast, Fidelity-oriented IQA is more suitable for traditional low-level image processing. Traditional low-level methods aim to remove degradation while preserving the original image content as much as possible. In this setting, image processing does not introduce new semantic information, and the high-level semantic content remains unchanged before and after processing \cite{TIP26sirWangAll}. Therefore, image quality can be mainly judged by how faithfully the processed image preserves the original content and structure, which makes Fidelity-oriented IQA suitable for traditional low-level image processing.

However, in recent years, deep learning has changed the paradigm of low-level image processing, especially with the development of large-scale data-driven methods and generative models such as GANs and diffusion models. Different from traditional methods, these approaches learn image patterns from large datasets rather than relying only on signal-level processing for each input image. To improve perceptual quality, they may introduce extra details during generation based on learned priors. This can produce semantically inconsistent content or hallucinated details \cite{TIP26sirWangAll, CVPR24sirDiffLiu, IJCV23sirXinDiffusion}, leading to local semantic changes. In other words, the semantic content of the processed image may differ from that of the input image from the perspective of machine perception. Existing IQA methods cannot properly reflect this kind of semantic distortion.

This issue is a fundamental one in low-level image processing. Low-level image processing aims to operate on image signals for tasks such as restoration, enhancement, reconstruction, and efficient transmission, while ideally preserving the semantic content of the original image. If semantic consistency is not maintained during processing, the output may contain semantic errors, missing objects, or incorrect relationships between objects. Semantic consistency evaluation focuses on whether the processed image preserves the original semantic content, without relying on any specific downstream task. Related ideas have already appeared in some application-specific settings. For example, in face super-resolution, some studies pay particular attention to whether the restored face preserves the identity-related information of the input \cite{TBBIS26iqaMachineChen, TBBIS26iqaFaceChen}. Although these methods are designed for a specific task, they suggest that image evaluation should consider not only visual fidelity, but also whether the original semantic content is preserved during processing.

Meanwhile, semantic similarity has already found applications in other domains. In semantic communication, some metrics are used to measure the consistency between transmitted images and received images \cite{Vitscore, CEMNLP21CLIPScore}. However, in this settings, image differences are mainly caused by channel noise and transmission distortion, rather than real changes in semantic structure. Because of this, such methods are still close to Fidelity-oriented IQA. They cannot effectively describe real semantic changes. Therefore, they do not provide a clear definition of Semantic Similarity as an independent evaluation task.

From the discussion above, it is clear that existing image quality evaluation methods cannot fully describe semantic changes between images. Therefore, image evaluation should include not only visual fidelity, but also semantic-level consistency. To address this need, we introduce a new evaluation task called \textit{Semantic Similarity}, which aims to measure semantic changes between images. We further provide a clear definition of this task, and discuss its applications and desired properties. We summarize our main contributions as follows.

First, we formalize \textit{Semantic Similarity} as a new evaluation task for low-level image processing, and provide a structured definition of image semantics together with the key properties and constraints of a valid semantic similarity index.

Second, we propose the first triplet-based semantic similarity index tailored for low-level image processing, termed Triplet-based Semantic Similarity Score (T3S), which jointly models foreground entities, background entities, and semantic relations.

Third, we conduct extensive experiments on COCO and SPA-Data, showing that T3S consistently outperforms representative baselines, better reflects progressive semantic changes, and that its key components are effective through ablation studies.

\section{Related Work}

\subsection{Existing IQA Methods}

Existing IQA methods can be broadly divided into three groups: Aesthetic-oriented IQA, Fidelity-oriented IQA, and Utility-oriented IQA.

Aesthetic-oriented IQA evaluates image quality mainly from the perspective of human visual preference. Recent works have improved this line of research by using semantic guidance, multimodal learning, multi-attribute supervision, and human feedback \cite{Aesthetic3, Aesthetic4, Aesthetic5, Aesthetic6, Aesthetic7, Aesthetic8}. However, these methods focus on whether an image looks pleasing to humans, rather than whether its semantic content is faithfully preserved during processing.

Fidelity-oriented IQA measures image quality by modeling distortion between images. Many studies in this area use CNN features, deep perceptual features, structural cues, or feature-domain statistics to better describe perceptual similarity \cite{Fidelity3, Fidelity4, Fidelity2, Fidelity5, Fidelity6, Fidelity7}. DeepSSIM \cite{Fidelity1} further improves robustness by modeling deep structural similarity. Although these methods are effective for traditional image restoration, they mainly focus on signal distortion and are less suitable for cases where semantic content may change during processing.

Utility-oriented IQA studies image quality from the perspective of downstream task usefulness. Existing works mainly evaluate whether an image is helpful for machine vision tasks such as detection or recognition \cite{Utility1, Utility2, Utility3}. This direction is related to semantic evaluation, but it is still different from our goal. Utility-oriented IQA depends on a specific downstream task, while our work focuses on whether the semantic content itself is preserved between images.

Overall, existing IQA methods evaluate image quality from visual preference, signal fidelity, or task usefulness. However, they do not explicitly define or measure semantic preservation and semantic change between images, which are important in modern low-level image processing.

\subsection{Semantic Evaluation in Communication}

Semantic similarity has also been explored in semantic communication. Most existing methods in this area measure similarity in a global feature space, instead of explicitly modeling the semantic structure of an image. For example, ViTScore \cite{Vitscore} uses ViT features to measure semantic similarity, and CLIPScore \cite{CEMNLP21CLIPScore} uses CLIP features for a similar purpose. SeSS \cite{Sess} further adds relation-aware information from scene graph generation to CLIP-based similarity, where relations are represented as triplets of \textit{subject}, \textit{predicate}, and \textit{object}.

These methods provide useful ideas for semantic-level assessment, but they still have some clear limits. Most of them measure semantic similarity mainly in a global feature space, so they are less effective in describing local semantic changes. Although SeSS introduces relation information, its core design still relies on global feature similarity and does not clearly define image semantics itself. In addition, these methods are mainly designed for semantic communication, where image differences usually come from transmission noise or distortion, rather than semantic changes introduced during image processing.

Therefore, existing IQA and semantic similarity methods provide useful inspiration, but they do not offer a clear framework for evaluating semantic preservation and semantic change in low-level image processing. This gap motivates us to explicitly define image semantics and build a new semantic similarity index for low-level image processing.


\section{Semantic Similarity in Low-Level Imaging}

\subsection{Definition of Semantic Similarity}


\begin{figure}[t] 
\centering 
\includegraphics[width=0.5\textwidth, keepaspectratio]{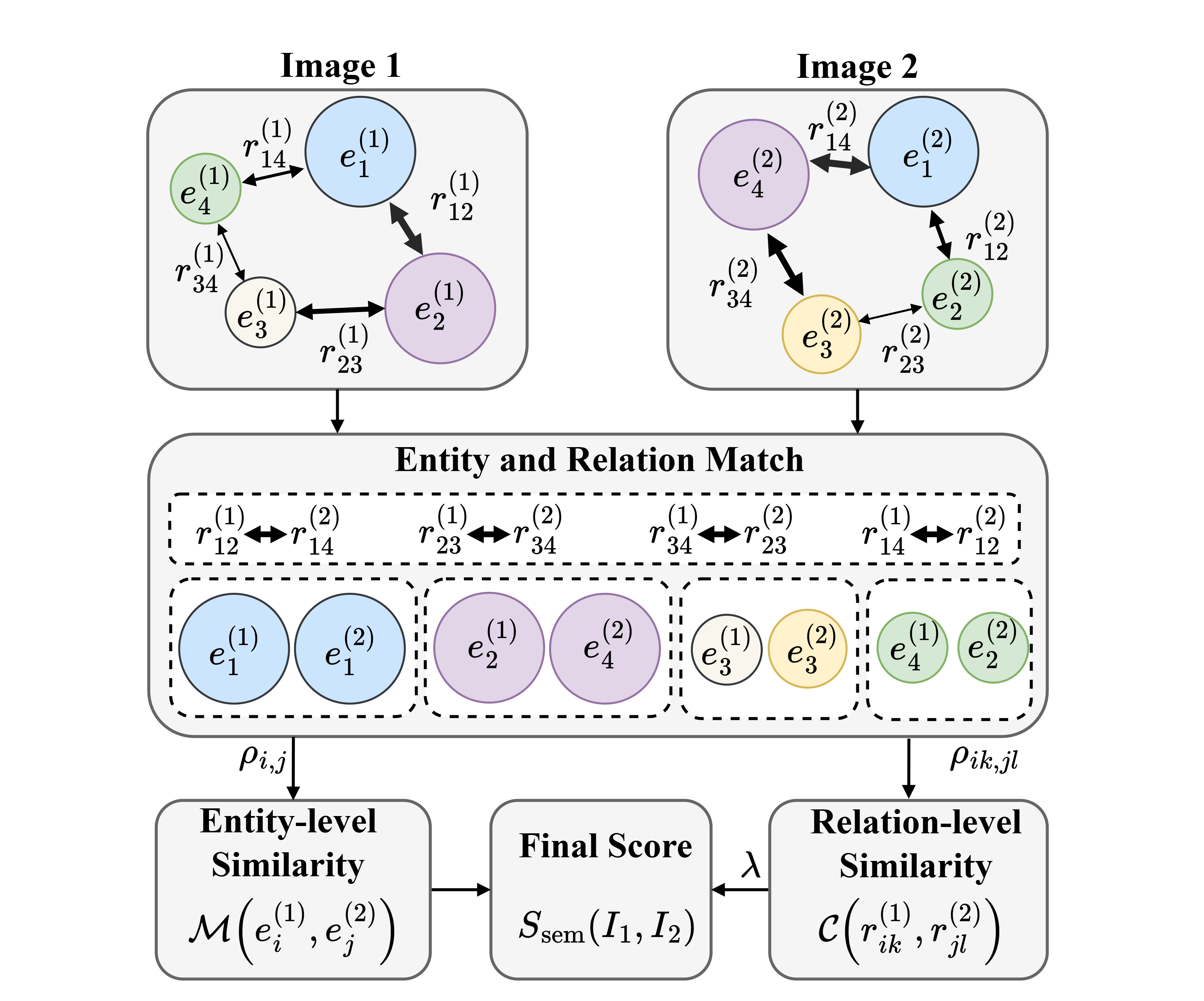} 
\caption{General formulation of semantic similarity based on pairwise weighted matching. The size of each entity denotes its semantic importance in computing $\rho_{i,j}$, while the relation links denote relation importance in computing $\rho_{ik,jl}$.} 
\label{fig:graph} 
\end{figure}

Existing studies in low-level processing, semantic communication, and related fields show that image semantics is usually described from four complementary aspects: semantic regions \cite{ACMMM20sieFanIntegrating, TIP22sieQiSGUIE, AAAI22sieLiangSemantically, TITS26sieJuSemantic, TIP26sirWangAll},  their category labels \cite{JSAC23sicTowardHuang}, global features \cite{vit, ICCV2021sirEmergingCaron, DINOv2, ICRA22sieWangSemantic, TPAMI25TaylorFormerJin, TMM24sirZhangPrior, TIP26sirZhangPerceive, Vitscore, CEMNLP21CLIPScore}, and relations \cite{Sess}.

Building on this observation, we introduce an entity-based paradigm for image semantic similarity assessment. Inspired by the notion of \emph{entity} in natural language processing \cite{TMM21entity}, we represent an image as a set of semantic entities (SEs), which serve as the basic units for describing image semantics for machine perception. Importantly, entities do not contribute equally to image semantics. Different entities may play different semantic roles and therefore should be weighted according to their semantic importance. Although existing studies often focus on semantic regions or global features, the unequal contribution of different entities has not been explicitly modeled. Accordingly, we define \textit{the semantics of an image} by \textit{its important entities---characterized by their category labels and local features---together with the relational information among them, including pairwise interactions and global scene context.}

Under this paradigm, an image $I$ is represented as
\begin{equation}
I=\{e_i\}_{i=1}^{N},
\end{equation}
where $e_i$ denotes a semantic entity and $N$ is the number of important entities in the image. Here, $e_i$ denotes an abstract semantic unit. In the practical implementation, its local visual feature and category semantics can be modeled separately and incorporated at different stages of the final similarity computation.

As shown in Fig.~\ref{fig:graph}, given two images $I_1$ and $I_2$, we define \textit{semantic similarity} as \textit{the degree to which their important entities and semantic relations are preserved across images.} Formally,
\begin{equation}
\begin{split}
S_{\text{sem}}(I_1,I_2)
&=
\sum_{i,j}\rho_{i,j}\mathcal{M}\!\left(e_i^{(1)}, e_j^{(2)}\right)  \\
&\quad+\lambda 
\sum_{(i,k),(j,l)}\rho_{ik,jl}\mathcal{C}\!\left(r_{ik}^{(1)}, r_{jl}^{(2)}\right),
\end{split}
\end{equation}
where $\rho_{i,j}$ denotes the importance weight assigned to each pair of matched entities $(e_i^{(1)}, e_j^{(2)})$ in the image pair $(I_1, I_2)$, where the matching of entities is established based on the similarity of the characteristics rather than the exact agreement of the category, and $\rho_{ik,jl}$ denotes the importance weight assigned to each pair of matched relations $(r_{ik}^{(1)}, r_{jl}^{(2)})$. The functions $\mathcal{M}(\cdot,\cdot)$ and $\mathcal{C}(\cdot,\cdot)$ measure the semantic correspondence of the entity-level and the relation-level, respectively, while $\lambda$ controls the relative contribution of the relation-level term. Moreover, category similarity can serve as one factor in determining $\rho_{i,j}$, so that entity pairs with more semantically similar categories may receive larger weights.

This formulation decomposes semantic similarity into two complementary parts: entity-level alignment and relation-level alignment. The first term measures semantic correspondence between entities in the two images, where $e_i^{(1)}$ and $e_j^{(2)}$ denote entities in $I_1$ and $I_2$, respectively, and $\rho_{i,j}$ captures the contribution of each entity pair. The second term measures the consistency of relations between entities, where $r_{ik}^{(1)}$ denotes the relation between $e_i^{(1)}$ and $e_k^{(1)}$ in $I_1$, and $r_{jl}^{(2)}$ is defined analogously for $I_2$; $\rho_{ik,jl}$ then captures the contribution of each relation pair. Overall, this formulation jointly models semantic entities and their relations in a pairwise weighted manner.

Overall, this formulation jointly models semantic entities and their relations, providing a more complete characterization of semantic similarity than conventional feature-based metrics.

\begin{figure*}[t] 
\centering 
\includegraphics[width=\textwidth, keepaspectratio]{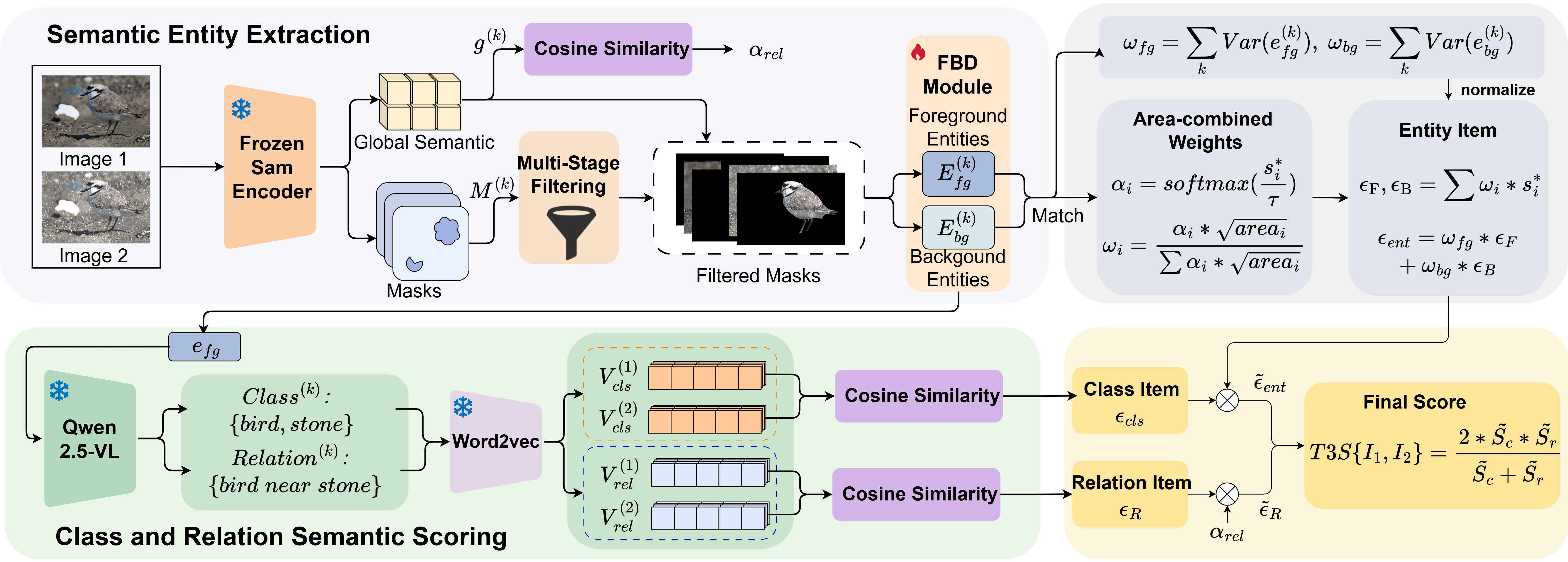} 
\caption{Overview of the proposed T3S framework. Given an input image pair, SAM first extracts SEs and decouples them into foreground and background to model the triplet components $\epsilon_{\mathrm{F}}$ and $\epsilon_{\mathrm{B}}$. Meanwhile, Qwen2.5-VL predicts class and relation labels from foreground entities to model the relational component $\epsilon_{\mathrm{R}}$. The final T3S index is obtained by coupling the triplet $(\epsilon_{\mathrm{F}}, \epsilon_{\mathrm{B}}, \epsilon_{\mathrm{R}})$.} 
\label{fig:T3S} 
\end{figure*}

Semantic similarity can be used as an evaluation index for both images and algorithms. For a single image, it measures whether the important content and relations in the reference image are preserved after processing. For an algorithm, it measures how well semantic content is preserved in its outputs, and thus provides a useful criterion for comparing different methods. At the current stage, Semantic similarity is not suitable as a training loss. Its computation depends on multiple discrete variables, such as entity extraction, category labeling, and relation prediction, which makes end-to-end differentiation difficult. Therefore, it is better suited for evaluation than for direct optimization.

\subsection{Constraints of Semantic Similarity}

For an image pair $I_1$ and $I_2$, we define a correspondence function $\pi(i)$ for entities in $I_1$, where $\pi(i)=j$ indicates that entity $e_i^{(1)}$ is matched to entity $e_j^{(2)}$, and $\pi(i)=0$ means that $e_i^{(1)}$ has no valid correspondence in $I_2$. In practice, this correspondence is established according to feature similarity between entities in the two images. Therefore, a matched entity pair does not necessarily have the same category label. A valid semantic similarity index should satisfy the following conditions.

\paragraph{Constraint 1: Invariance to non-semantic distortions.}

Under an ideal low-level processing setting, if two images differ only in pixel appearance but their SEs and relations remain unchanged under machine perception, the semantic similarity should remain high. In this case, all entities are assumed to have valid one-to-one correspondences, and their semantic categories and relations should be preserved:
\begin{equation}
c\!\left(e_i^{(1)}\right)=c\!\left(e_{\pi(i)}^{(2)}\right),\;
r_{ik}^{(1)}=r_{\pi(i)\pi(k)}^{(2)},
\quad \forall\, i\neq k .
\end{equation}
Here, $c(e)$ denotes the semantic category label of entity $e$. This corresponds to the ideal case where SE extraction is correct and semantic content is fully preserved before and after low-level processing. Therefore, $S_{\text{sem}}(I_1,I_2)$ should approach its upper bound:
\begin{equation}
S_{\text{sem}}(I_1,I_2)\rightarrow 1.
\end{equation}

\paragraph{Constraint 2: Sensitivity to entity-level semantic shift.}

If low-level processing changes the semantic identity of one or more entities, the similarity index should decrease accordingly. Let $\Omega_E$ be the set of shifted entities:
\begin{equation}
\Omega_E=\left\{ i \;\middle|\; \pi(i)=0 \ \text{or}\; c\!\left(e_i^{(1)}\right) \neq c\!\left(e_{\pi(i)}^{(2)}\right) \right\}.
\end{equation}
Then $S_{\text{sem}}(I_1,I_2)$ should decrease as the number of shifted entities increases. The penalty assigned to each shifted entity $i\in\Omega_E$ should depend on $\rho_i, \rho_j$, such that entities with larger weights receive stronger penalties.

\paragraph{Constraint 3: Sensitivity to relation-level semantic shift.}

Even when the entities themselves are preserved, the similarity index should decrease if the relations between entities change. Let $\Omega_R$ be the set of shifted relations:
\begin{equation}
\Omega_R=\left\{(i,k)\;\middle|\; \pi(i)=0 \ \text{or} \ \pi(k)=0 \ \text{or} \ r_{ik}^{(1)} \neq r_{\pi(i)\pi(k)}^{(2)}\right\}.
\end{equation}
Then $S_{\text{sem}}(I_1,I_2)$ should decrease as the number of shifted relations increases. The penalty assigned to each shifted relation $(i,k)\in\Omega_R$ should depend on the importance of the entities involved, so that relations between more important entities incur larger penalties.

\paragraph{Constraint 4: Normalized score range.}
The semantic similarity metric should be bounded in the range $[-1,1]$:
\begin{equation}
S_{\text{sem}}(I_1,I_2)\in[-1,1].
\end{equation}
Here, $S_{\text{sem}}=1$ denotes perfect positive semantic correlation, $S_{\text{sem}}=0$ denotes no semantic correlation, and $S_{\text{sem}}=-1$ denotes perfect negative semantic correlation. Similar to the phenomenon observed in \cite{Vitscore}, the score is typically not smaller than 0 in practice, because most image pairs are at most semantically unrelated rather than negatively correlated.

A valid semantic similarity index should satisfy the above constraints, remain stable under non-semantic changes while being sensitive to real entity-level and relation-level semantic shifts, and provide an interpretable, continuous, and unified measure across invariance to non-semantic perturbations.

\section{Proposed T3S Model}

From the above analysis, image Semantic Similarity is jointly determined by semantic entities (SEs) and their relations, and thus cannot be adequately described by a single global representation. Inspired by the structured modeling philosophy of E-CARGO \cite{ecargo}, we argue that image semantics should be explicitly organized as a structured system before quantitative evaluation. The central idea of E-CARGO is that a complex system should first be described through its basic units and their organizational relations, and only then subjected to quantitative analysis.

Following this perspective, we reorganize image semantics according to the structure of semantic perception. Prior work has shown that foreground and background do not contribute equally to image semantics \cite{AAAI22sieLiangSemantically}. Motivated by this observation, we decompose image semantics into three components---foreground, background, and relation---and define the Triplet-based Semantic Similarity Score (T3S). For an image pair $(I_1,I_2)$, T3S is formulated as
\begin{equation}
T3S(I_1,I_2)=\Phi\!\left(\epsilon_{\text{F}},\epsilon_{\text{B}},\epsilon_{\text{R}}\right),
\end{equation}
where $\epsilon_{\text{F}}$, $\epsilon_{\text{B}}$, and $\epsilon_{\text{R}}$ denote the discrepancies in foreground entities, background entities, and semantic relations, respectively, and $\Phi(\cdot)$ maps these triplet components to the final similarity score. Here, $\epsilon_{\text{F}}$ and $\epsilon_{\text{B}}$ jointly characterize entity-level variation: $\epsilon_{\text{F}}$ is derived from the features and category labels of foreground entities, whereas $\epsilon_{\text{B}}$ measures background differences. $\epsilon_{\text{R}}$ captures relation-level variation through $\langle \textit{subject}, \textit{predicate}, \textit{object} \rangle$ triplets.

\subsection{Semantic Entity Extraction and Scoring}

Given an input image pair $(I_1, I_2)$, this part extracts SEs and measures their entity-level consistency. Specifically, we use the pre-trained \textit{Segment Anything Model} (SAM) to generate candidate SEs. For each image $I_k$, SAM first generates a set of candidate regions
\begin{equation}
M^{(k)}=\{m_i^{(k)}\}, \qquad k\in\{1,2\}.
\end{equation}
We then apply a multi-stage filtering strategy to obtain reliable SEs. Specifically, candidates with overly small regions or excessively large coverage are first removed based on their mask areas. The remaining candidates are ranked by
\begin{equation}
q_i^{(k)}=\mathrm{predicted\_iou}\!\left(m_i^{(k)}\right)\cdot \mathrm{stability\_score}\!\left(m_i^{(k)}\right),
\end{equation}
and redundant overlapping masks are further removed by deduplication. The retained regions are taken as the core SEs.

For each retained SE, we extract its feature from the SAM encoder feature map, denoted as $e_i^{(k)} \in \mathbb{R}^d$. Meanwhile, we directly extract an image-level feature from the intermediate output of the encoder, $g^{(k)} \in \mathbb{R}^d$, to represent the global semantic information of the whole image. Based on this, the modulation weight for the relation term is defined as
\begin{equation}
\alpha_{rel} =
\frac{g^{(1)\top} g^{(2)}}{\|g^{(1)}\|_2 \, \|g^{(2)}\|_2}.
\end{equation}
Here, $\alpha_{rel}$ is used as a global semantic prior for relation modeling.

To separate retained semantic entities into foreground and background groups, we propose a Foreground--Background Decoupling (FBD) module. The key idea is that region area can serve as one cue for entity importance, while background regions are often large but semantically less informative. FBD first disentangles background entities from the retained set, so that later weighting can be applied more appropriately to foreground entities. It uses an attraction term to enhance foreground coherence and a repulsion term to suppress background interference by increasing feature separation. Since the same decomposition is applied independently to both input images, we describe the procedure for one image below. Given the retained entity set $E=\{{e_i}\}_{i=1}^{N}$, we first project each entity feature into a higher-dimensional latent space:
\begin{equation}
z_i^{(0)} = W_e e_i + b_e .
\end{equation}
Here, $e_i$ denotes the visual feature of the $i$-th retained semantic entity extracted from SAM. Its category label is not included at this stage and will be predicted separately in Sec.~4.2.

\begin{figure}[t] 
\centering 
\includegraphics[width=0.5\textwidth, keepaspectratio]{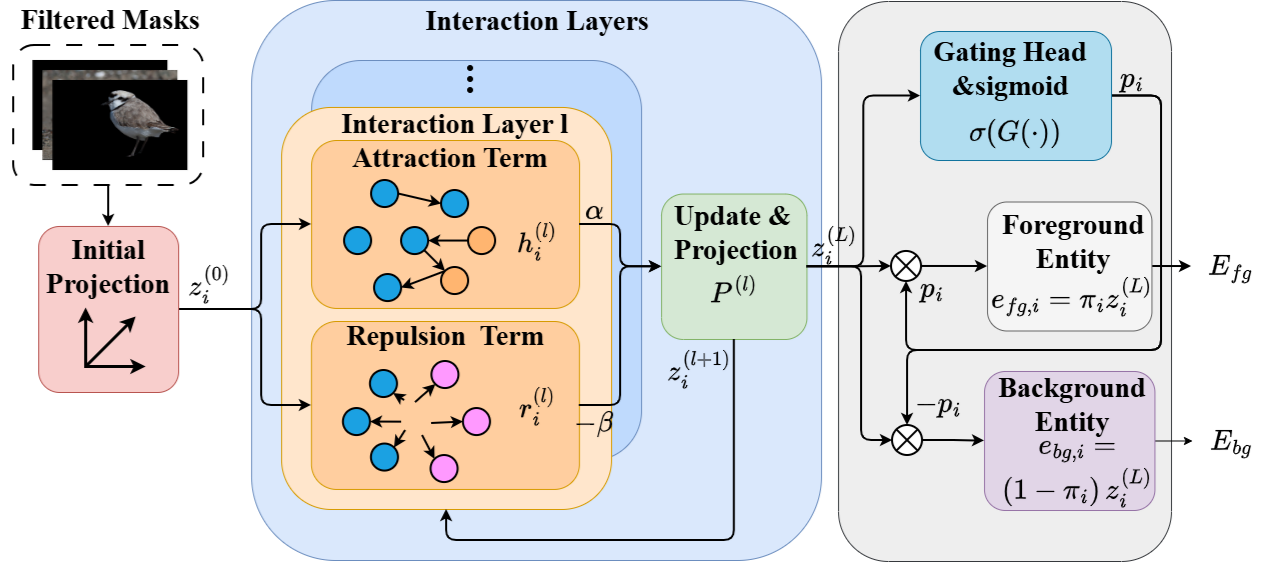} 
\caption{The FBD module performs feature disentanglement via attraction and repulsion. It pulls semantically consistent features together and pushes inconsistent ones apart, thereby enhancing the distinction between foreground and background entities.} 
\label{fig:fbdm} 
\end{figure}

The resulting feature set is then refined by interaction layers. For the $l$-th layer, let
\begin{equation}
d_{ij}^{(l)} = \| z_i^{(l)} - z_j^{(l)} \|_2
\end{equation}
denote the pairwise distance between two entities. Based on this, an attraction term is defined as
\begin{equation}
w_{ij}^{(l)} = \mathrm{softmax}_j \!\left( - (d_{ij}^{(l)})^2 \right), \qquad
h_i^{(l)} = \sum_j w_{ij}^{(l)} z_j^{(l)},
\end{equation}
which aggregates semantically coherent entities in the feature space. Meanwhile, a repulsion term is introduced as
\begin{equation}
r_i^{(l)} = \sum_j \exp\!\left(-d_{ij}^{(l)}\right)\left(z_i^{(l)} - z_j^{(l)}\right),
\end{equation}
which enlarges the separation between ambiguous entities. The refined feature is then updated by
\begin{equation}
z_i^{(l+1)} = P^{(l)}\!\left(z_i^{(l)} + \alpha h_i^{(l)} - \beta r_i^{(l)}\right),
\end{equation}
where $\alpha$ and $\beta$ control the strengths of attraction and repulsion, respectively, and $P^{(l)}(\cdot)$ denotes a learnable linear projection.

After the two-stage field interaction, a lightweight gating head produces a foreground partition coefficient for each entity:
\begin{equation}
p_i = \sigma\!\left(G\!\left(z_i^{(L)}\right)\right),
\end{equation}
where $p_i \in [0,1]$ measures the degree to which the $i$-th entity belongs to the primary semantic component, $\sigma(\cdot)$ denotes the sigmoid function, $G(\cdot)$ is a lightweight gating head, and $L$ is the total number of layers. Since the same decomposition is performed for each input image, we restore the image index $k \in \{1,2\}$ and define the foreground and background entities as
\begin{equation}
e_{fg,i}^{(k)} = p_i^{(k)} z_i^{(L,k)}, \qquad
e_{bg,i}^{(k)} = \left(1-p_i^{(k)}\right) z_i^{(L,k)}.
\end{equation}
Accordingly, the retained entity set is softly decomposed into foreground and background components:
\begin{equation}
E_{fg}^{(k)} = \left\{ e_{fg,i}^{(k)} \right\}_{i=1}^{N_k}, \qquad
E_{bg}^{(k)} = \left\{ e_{bg,i}^{(k)} \right\}_{i=1}^{N_k}.
\end{equation}

Based on the foreground and background entity sets from two images, i.e.,
$(E_{fg}^{(1)}, E_{fg}^{(2)})$ and $(E_{bg}^{(1)}, E_{bg}^{(2)})$,
we further compute the corresponding matching scores. For a given entity-set pair,
denoted generically as $E^{(1)}=\{e_i^{(1)}\}$ and $E^{(2)}=\{e_j^{(2)}\}$,
we first identify, for each entity $e_i^{(1)}$, the most similar entity in $E^{(2)}$ by cosine similarity:
\begin{equation}
j^*(i)=\arg\max_j \frac{(e_i^{(1)})^\top e_j^{(2)}}{\|e_i^{(1)}\|_2 \, \|e_j^{(2)}\|_2}.
\end{equation}
The resulting similarity value is denoted by $s_i^*$, representing the cosine similarity between $e_i^{(1)}$ and its best-matched counterpart $e_{j^*(i)}^{(2)}$.

We further assign normalized weights according to their matching confidence:
\begin{equation}
\alpha_i =
\frac{\exp(s_i^*/\tau)}
{\sum_{t \in \Omega} \exp(s_t^*/\tau)},
\end{equation}
where $\Omega$ denotes the set of valid matches and $\tau$ is the temperature parameter. The final weight of each matched entity is defined as
\begin{equation}
\tilde{w}_i =
\frac{\alpha_i\cdot area_i^{(1)}}
{\sum_{t \in \Omega} \alpha_t\cdot area_t^{(1)}}.
\end{equation}
Here, $area_i^{(1)}$ denotes the mask area of the $i$-th matched entity in image $I_1$. The weight $\tilde{w}_i$ is determined jointly by the matching confidence $\alpha_i$ and the entity area $area_i^{(1)}$. The matching score between two entity sets is then defined as:
\begin{equation}
S(E_1,E_2) = \sum_{i \in \Omega} \tilde{w}_i s_i^*.
\end{equation}

Accordingly, the foreground and background matching scores are defined as
\begin{equation}
\epsilon_{\text{F}} = S(E_{fg}^{(1)}, E_{fg}^{(2)}), \qquad
\epsilon_{\text{B}} = S(E_{bg}^{(1)}, E_{bg}^{(2)}).
\end{equation}

To quantify their relative importance, we estimate the smoothness of foreground and background entities using feature variance and use it to adaptively adjust their scores:
\begin{equation}
v_{fg} = \mathrm{Var}(E_1^{fg}) + \mathrm{Var}(E_2^{fg}), \qquad
v_{bg} = \mathrm{Var}(E_1^{bg}) + \mathrm{Var}(E_2^{bg}),
\end{equation}
where $\mathrm{Var}(\cdot)$ denotes the feature variance. The initial weights are defined as
\begin{equation}
w_{fg}' = \frac{v_{fg}}{v_{fg} + \lambda_{bg} v_{bg}}, \qquad
w_{bg}' = \frac{\lambda_{bg} v_{bg}}{v_{fg} + \lambda v_{bg}},
\end{equation}
where $\lambda_{bg}$ is a background attenuation factor used to suppress the contribution of background entities. After normalization, the final fusion coefficients are obtained as $w_{fg}$ and $w_{bg}$, respectively.

The final entity-level score is defined as:
\begin{equation}
 \epsilon_{\mathrm{ent}} = w_{fg} \epsilon_{\text{F}} + w_{bg} \epsilon_{\text{B}}.
\end{equation}

\subsection{Class and Relation Semantic Scoring}

We use Qwen2.5-VL 3B as the extractor of class and relation labels. Specifically, given an input image pair $(I_1, I_2)$, we feed the foreground entity sets $E_{fg}^{(1)}$ and $E_{fg}^{(2)}$ into the model, and use a constrained prompt to extract main object classes and their explicit relations:

\begin{equation}
Class^{(k)} = \{ c_i^{(k)} \}, \ \ \ 
Relation^{(k)} = \{ (c_u^{(k)}, p_{uv}^{(k)}, c_v^{(k)}) \}. \qquad
\end{equation}
Here, $Class^{(k)}$ denotes the class label set of the $k$-th image, and $Relation^{(k)}$ denotes its relation label set. $c_i^{(k)}$ denotes the class label of the $i$-th entity, and $(c_u^{(k)}, p_{uv}^{(k)}, c_v^{(k)})$ denotes a relation triplet.

To compare object-level semantic consistency between two images, we first convert the object lists into token sequences and embed them using pre-trained Word2Vec. Let $\phi(\cdot)$ denote the pre-trained Word2Vec embedding function. The class and relation embedding sets of the $k$-th image are defined as
\begin{equation}
\begin{aligned}
\mathbf{V}_{cls}^{(k)} &= \left\{ \phi\!\left(c_i^{(k)}\right) \right\}_{i=1}^{N_k}, \\
\mathbf{V}_{rel}^{(k)} &= \left\{ \left[ \phi\!\left(c_u^{(k)}\right) ; \phi\!\left(p_{uv}^{(k)}\right) ; \phi\!\left(c_v^{(k)}\right) \right] \right\}. \\
\end{aligned}
\end{equation}
Here, $k \in \{1,2\}$ and ${(c_u^{(k)},\,p_{uv}^{(k)},\,c_v^{(k)}) \in Relation^{(k)}}$, $\mathbf{V}_{cls}^{(k)}$ denotes the class embeddings of $E_{fg}^{(k)}$, where $N_k$ is the number of labels. $\mathbf{V}_{rel}^{(k)}$ denotes relation embeddings. $[\cdot;\cdot;\cdot]$ denotes vector concatenation.

After zero-padding the shorter sequences, we compute the similarity matrices for the class and relation embeddings:
\begin{equation}
\epsilon_{ij}^{cls} =
\frac{\left(\mathbf{v}_{cls,i}^{(1)}\right)^\top \mathbf{v}_{cls,j}^{(2)}}
{\left\|\mathbf{v}_{cls,i}^{(1)}\right\|_2 \, \left\|\mathbf{v}_{cls,j}^{(2)}\right\|_2},
\qquad
\epsilon_{ij}^{rel} =
\frac{\left(\mathbf{v}_{rel,i}^{(1)}\right)^\top \mathbf{v}_{rel,j}^{(2)}}
{\left\|\mathbf{v}_{rel,i}^{(1)}\right\|_2 \, \left\|\mathbf{v}_{rel,j}^{(2)}\right\|_2}.
\end{equation}

The class-level and relation-level semantic consistency are then measured by the average of the maximum matching similarities:

\begin{equation}
\alpha_{\text{cls}} = \frac{1}{N}\sum_i \max_j s_{ij}^{cls}, \qquad
\epsilon_{\text{R}} = \frac{1}{M}\sum_i \max_j s_{ij}^{rel}.
\end{equation}

\subsection{Overall T3S Formulation}

We augment the entity term with class consistency and weight the relation term by the global semantic factor $\alpha_{rel}$:
\begin{equation}
\tilde{\epsilon}_{\mathrm{ent}} = \epsilon_{\mathrm{ent}} \cdot \alpha_{\mathrm{cls}}, \qquad
\tilde{\epsilon}_{\mathrm{R}} = \alpha_{rel} \cdot \epsilon_{\mathrm{R}}.
\end{equation}
We then use the harmonic mean of the object term and the relation term as the final T3S index:
\begin{equation}
T3S(I_1, I_2) =
\frac{2 \tilde  \epsilon_{\mathrm{ent}} \tilde \epsilon_{\mathrm{R}}}
{\tilde  \epsilon_{\mathrm{ent}} + \tilde \epsilon_{\mathrm{R}}}.
\end{equation}
Therefore, the final index is obtained only when both entity and relation semantics are well preserved. Obviously, the proposed T3S has successfully fulfilled the four constrains mentioned in Section 3.2. {\it Details of the discussions are included in the supplementary document. }

\begin{table*}[t]
\centering
\caption{Average scores on COCO and SPA-Data over five degradation levels. For each column, the highest score is highlighted in red and the lowest score is highlighted in blue.}
\label{tab:avg_results}
\small
\begin{tabular*}{0.9\textwidth}{@{\extracolsep{\fill}}lcccccccccc}
\toprule
& \multicolumn{5}{c}{COCO} & \multicolumn{5}{c}{SPA-Data} \\
\cmidrule(lr){2-6} \cmidrule(lr){7-11}
Method & Derain & Desnow & Deblur & Denoise & Overall & Derain & Desnow & Deblur & Denoise & Overall \\
\midrule
SeSS      & 0.7837 & 0.6617 & 0.5846 & 0.5947 & 0.6562 & \textcolor{blue}{0.7533} & 0.6389 & 0.6541 & 0.5688 & 0.6538 \\
DeepSSIM  & \textcolor{blue}{0.7720} & 0.4904 & \textcolor{blue}{0.4290} & \textcolor{blue}{0.4912} & 0.5456 & 0.7941 & 0.4781 & \textcolor{blue}{0.5484} & \textcolor{blue}{0.4369} & \textcolor{blue}{0.5644} \\
SSIM      & 0.8376 & 0.5844 & 0.5908 & \textcolor{blue}{0.3313} & 0.5860 & 0.8499 & \textcolor{blue}{0.4740} & 0.8030 & 0.2385 & 0.5913 \\
ViTScore  & 0.8241 & \textcolor{blue}{0.4638} & 0.7230 & 0.8026 & 0.7034 & 0.7775 & 0.4833 & 0.7779 & 0.7580 & 0.6992 \\
T3S(ours) & \textcolor{red}{\textbf{0.9080}} & \textcolor{red}{\textbf{0.8154}} & \textcolor{red}{\textbf{0.8686}} & \textcolor{red}{\textbf{0.8763}} & \textcolor{red}{\textbf{0.8671}}
          & \textcolor{red}{\textbf{0.9008}} & \textcolor{red}{\textbf{0.7824}} & \textcolor{red}{\textbf{0.8871}} & \textcolor{red}{\textbf{0.8328}} & \textcolor{red}{\textbf{0.8508}} \\
\bottomrule
\end{tabular*}
\end{table*}

\begin{figure}[t] 
\centering 
\includegraphics[width=0.5\textwidth, keepaspectratio]{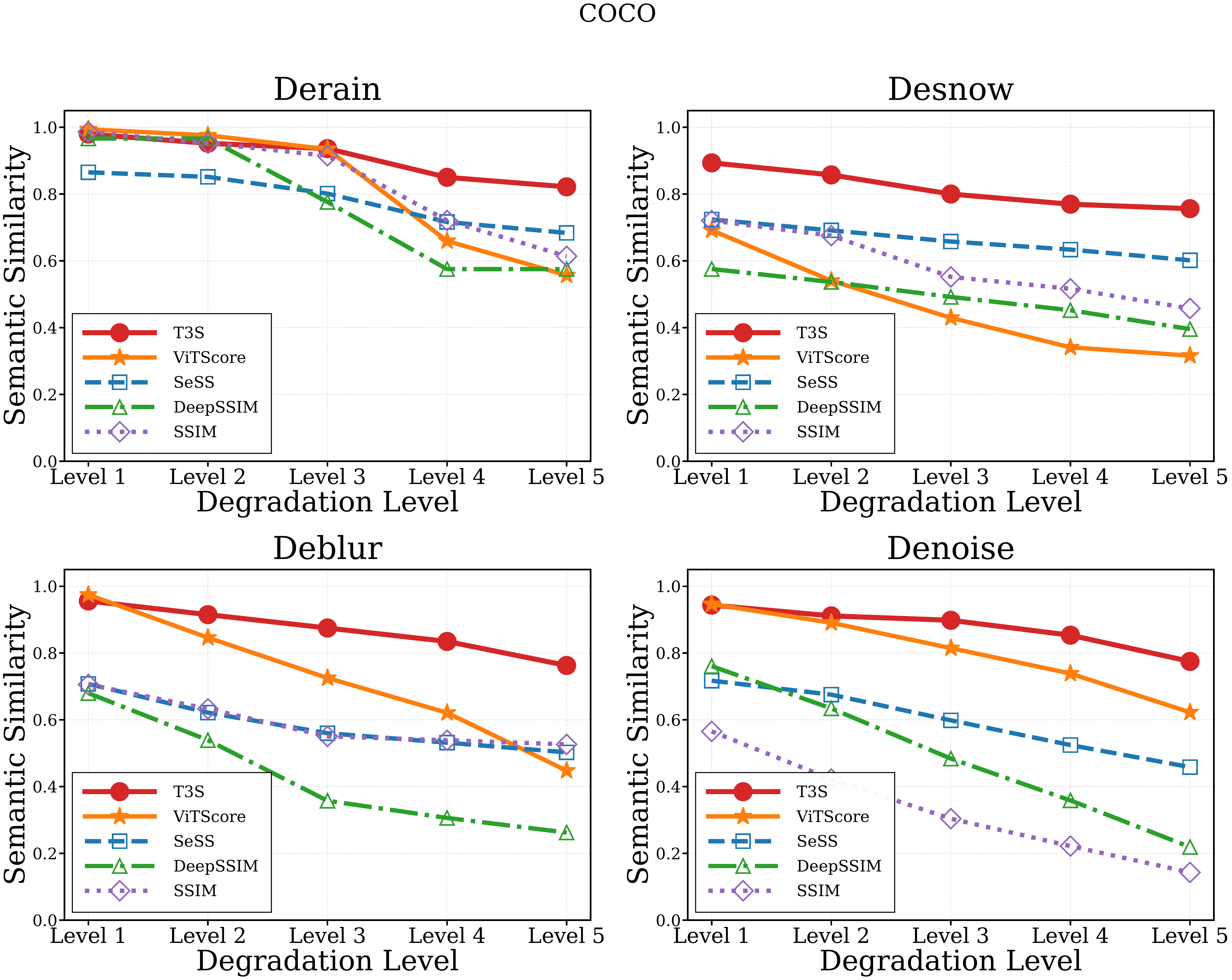} 
\caption{Comparison of score curves of different methods on the COCO dataset under four low-level image processing tasks.} 
\label{fig:coco_curves} 
\end{figure}

\begin{figure}[t] 
\centering 
\includegraphics[width=0.5\textwidth, keepaspectratio]{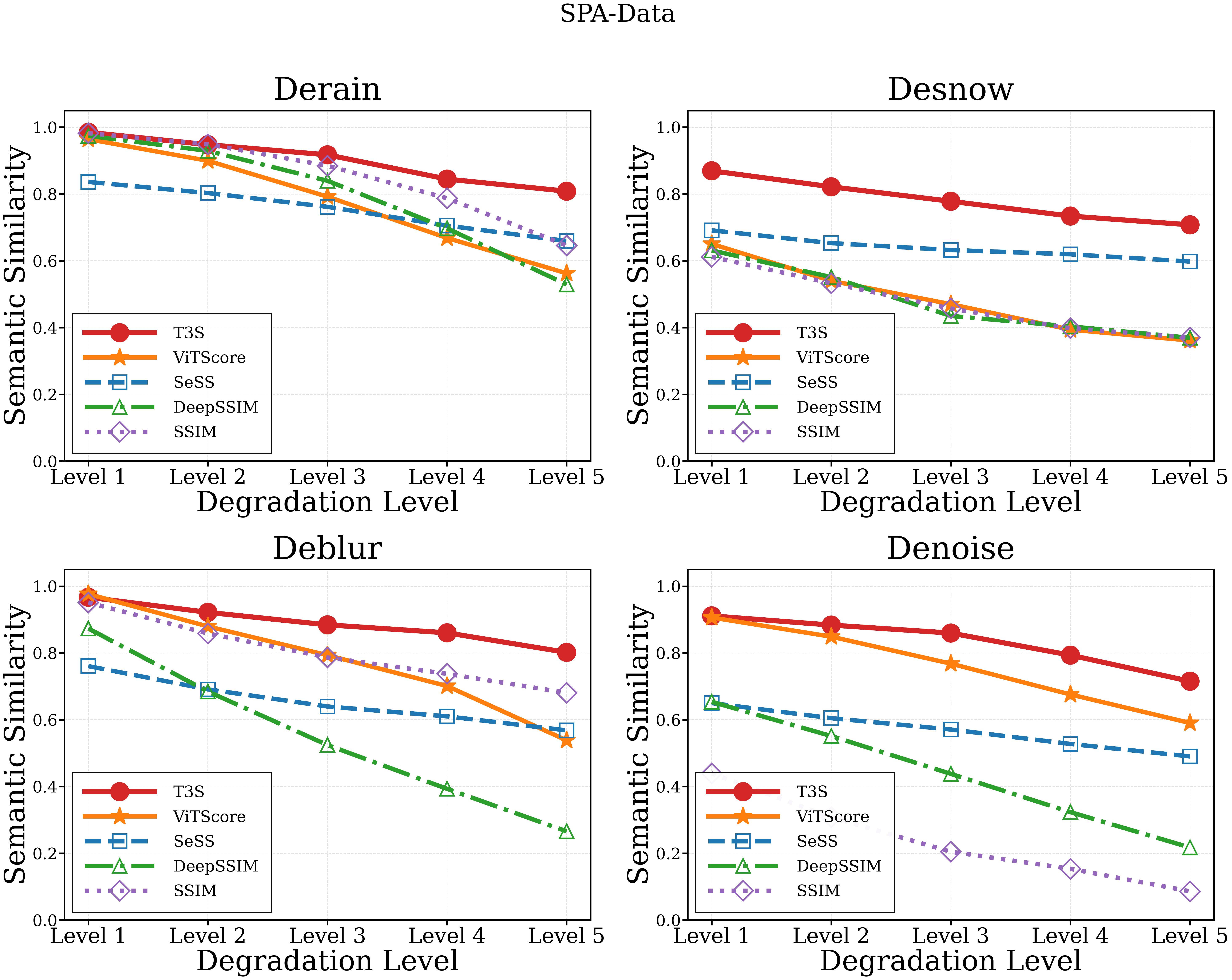} 
\caption{Comparison of score curves of different methods on the SPA-Data dataset under four low-level image processing tasks.} 
\label{fig:spa_curves} 
\end{figure}

\section{Experiments}

\subsection{Experimental Setup}

\textbf{Databases.} We conduct experiments on the SPA-Data and COCO \cite{coco} datasets. For each dataset, we randomly sample 1,000 ground-truth images and use the open-source degradation pipeline from \cite{HiKER} to synthesize 20 common degradation types, each with 5 severity levels. This setting provides a diverse benchmark for semantic similarity assessment under low-level degradations. Due to space limitations, the main paper reports results on four representative tasks---deraining, desnowing, denoising, and deblurring---while the complete results are provided in the supplementary material.

\textbf{Tasks and Metrics.} We evaluate the proposed method by measuring semantic similarity between each degraded image and its ground-truth image. Based on the constraints in Section 3, a valid semantic similarity index should remain high when semantic content is preserved and decrease as degradation severity or semantic shift increases. We therefore focus on two criteria: stability under degradations and sensitivity to increasing degradation severity. We report average similarity scores across degradation types and levels, and further examine whether each index exhibits a clear and consistent decline as degradation becomes stronger.

\begin{figure*}[t] 
\centering 
\includegraphics[width=\textwidth, keepaspectratio]{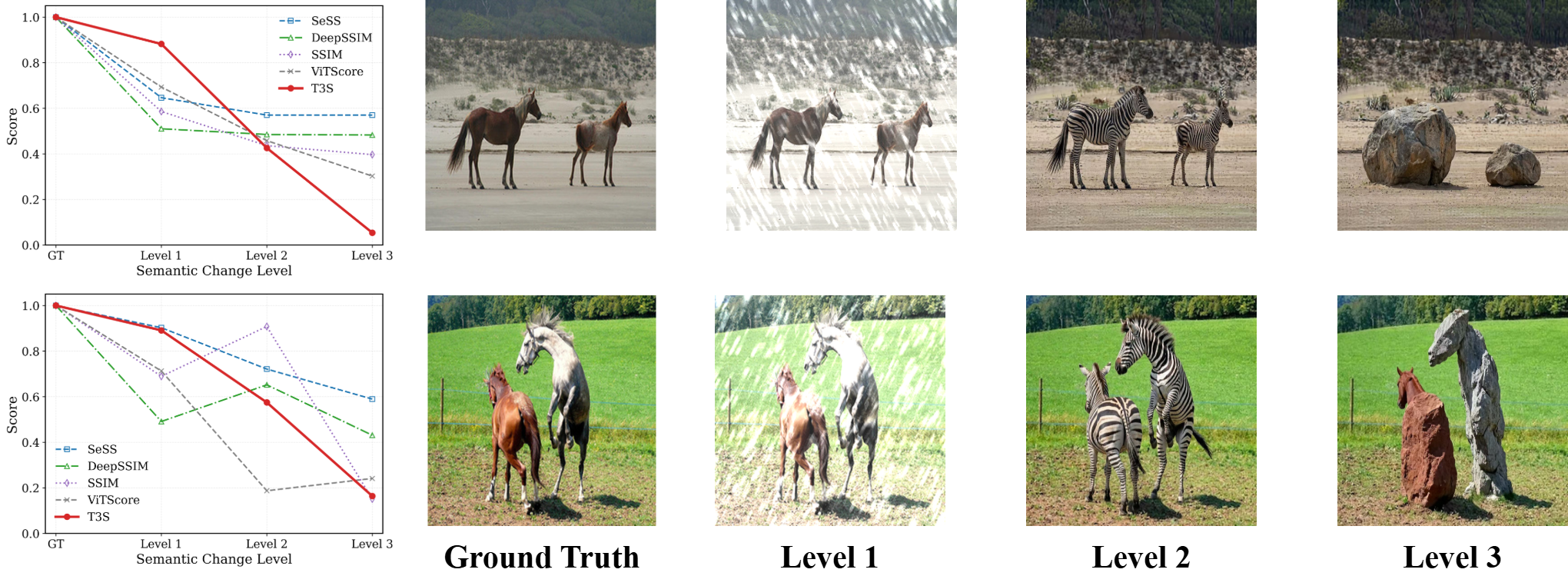} 
\caption{This figure summarizes our evaluation protocol under different levels of semantic change. To assess the proposed method under varying degrees of semantic change, we construct multiple levels of semantic variation through image processing and perform experiments accordingly.} 
\label{fig:changesemantic} 
\end{figure*}

\textbf{Implementation Details.} Our method is implemented in PyTorch and all experiments are conducted on a single NVIDIA RTX 4090 GPU. We use the pre-trained SAM-H model for candidate mask generation. During preprocessing, low-quality candidate masks are filtered out. For category and relation extraction, we use Qwen2.5-VL-3B. For the FBD module, we annotate SAM outputs on COCO as foreground or background and train the module separately with a binary classification loss.

\textbf{Baselines.} We compare proposed method with four representative metrics: two fidelity-oriented IQA methods, \textbf{SSIM} and \textbf{DeepSSIM}, and two semantic similarity methods from semantic communication, \textbf{ViTScore} and \textbf{SeSS}. First group emphasizes fidelity and structural consistency, while the second measures semantic consistency mainly through high-level feature alignment. This comparison enables us to assess the proposed method against both traditional IQA baselines and representative semantic similarity baselines.

\subsection{Results and Discussions}

From Table~\ref{tab:avg_results}, we can observe that T3S achieves the best results in all compared settings on both COCO and SPA-Data. Compared with existing metrics, the proposed method shows clear and stable advantages not only in overall average score, but also in each individual task. This suggests that T3S can better preserve evaluation consistency under different degradation types. In particular, its larger margins on desnowing and deblurring indicate that structured semantic modeling is especially beneficial when degradations heavily interfere with scene content and object relationships. These results demonstrate the effectiveness of proposed triplet-based formulation for semantic similarity assessment in low-level image processing.

As shown in Figures~\ref{fig:coco_curves} and \ref{fig:spa_curves}, all methods show decreasing scores as degradation level increases, but T3S degrades much more slowly than baselines. Its advantage is relatively small under mild distortions, yet becomes increasingly clear in medium-to-severe regime. This pattern is consistent across both COCO and SPA-Data and across all four tasks. In particular, T3S successfully maintains the highest scores under severe deraining, desnowing, deblurring, and denoising, indicating stronger robustness to semantic corruption caused by heavy degradation. These results suggest that the proposed triplet-based modeling provides a more stable and reliable measure of semantic similarity than existing fidelity-based or feature-based metrics.

Fig.~\ref{fig:changesemantic} shows the responses of different metrics when GT image is compared with three levels of semantically changed images. Level 1 adds rain streaks to the GT image and causes only a small semantic change. Level 2 replaces horse with zebra, leading to a moderate semantic change because the two subjects are still close in category. Level 3 replaces horse with stone, causing a much larger semantic change because the new subject is far from the original category in meaning. In both examples, T3S decreases monotonically from 0.8821/0.8910 to 0.4259/0.5747 and further to 0.0539/0.1640, whereas the other metrics show less consistent trends. These results indicate that T3S is capable of better reflecting the degree of semantic change.

\begin{table}[t]
\centering
\caption{Ablation study on low-level image processing tasks.}
\label{tab:ablation}
\setlength{\tabcolsep}{5pt}
\small
\begin{tabular}{cc|cccc}
\toprule
FBD & Rel. & Derain & Desnow & Deblur & Denoise \\
\midrule
$\checkmark$ & $\checkmark$ & \textbf{0.9079} & \textbf{0.8154} & \textbf{0.8686} & \textbf{0.8737} \\
$\times$     & $\checkmark$ & 0.8794 & 0.7882 & 0.8562 & 0.8593 \\
$\checkmark$ & $\times$     & 0.8375 & 0.6531 & 0.7738 & 0.7799 \\
$\times$     & $\times$     & 0.7926 & 0.6417 & 0.7569 & 0.7594 \\
\bottomrule
\end{tabular}
\end{table}

\subsection{Ablation Studies}

Table~\ref{tab:ablation} summarizes the ablation results over five degradation levels for four low-level image processing tasks. In table, FBD refers to the proposed Foreground--Background Decoupling module, and \textit{Rel.} refers to the relation branch. The complete model achieves the best performance on all tasks, with scores of 0.9079, 0.8154, 0.8686, and 0.8737 on deraining, desnowing, deblurring, and denoising, respectively.

The variant \textit{w/o FBD} shows consistent declines across all these tasks, confirming the benefit of foreground-background disentanglement. This effect is more evident in deraining and desnowing, where degradations introduce stronger background interference. In comparison, \textit{w/o Rel.} causes a larger drop overall, highlighting the importance of relational information for semantic consistency modeling; this effect is especially clear in desnowing, where the score falls from 0.8154 to 0.6531. Finally, \textit{w/o FBD \& Rel.} produces the worst performance on all the four tasks. These results indicate that FBD and the relation branch are complementary, and that both are necessary for a more complete characterization of semantic preservation.

\section{Conclusion}

In this paper, we present Semantic Similarity as a new evaluation perspective for low-level image processing by explicitly specifying what semantic preservation should mean in this context. To this end, we represent image semantics through semantic entities and their structural relations, and systematically discuss the functional role, core constraints, and desirable properties of semantic similarity as an independent evaluation problem. Building on this formulation, we introduce T3S as a concrete metric that instantiates the proposed perspective. Extensive experiments on COCO and SPA-Data demonstrate that T3S consistently outperforms existing methods and aligns more closely with progressive semantic changes across diverse degradations. Taken together, these results underscore the importance of semantic-aware evaluation in modern low-level vision and lay a conceptual basis for future work on semantics-oriented assessment.

\bibliographystyle{ACM-Reference-Format}
\bibliography{references}

\clearpage
\appendix

\section{Additional Experimental Results}

\begin{strip}
\centering
\includegraphics[width=\textwidth, keepaspectratio]{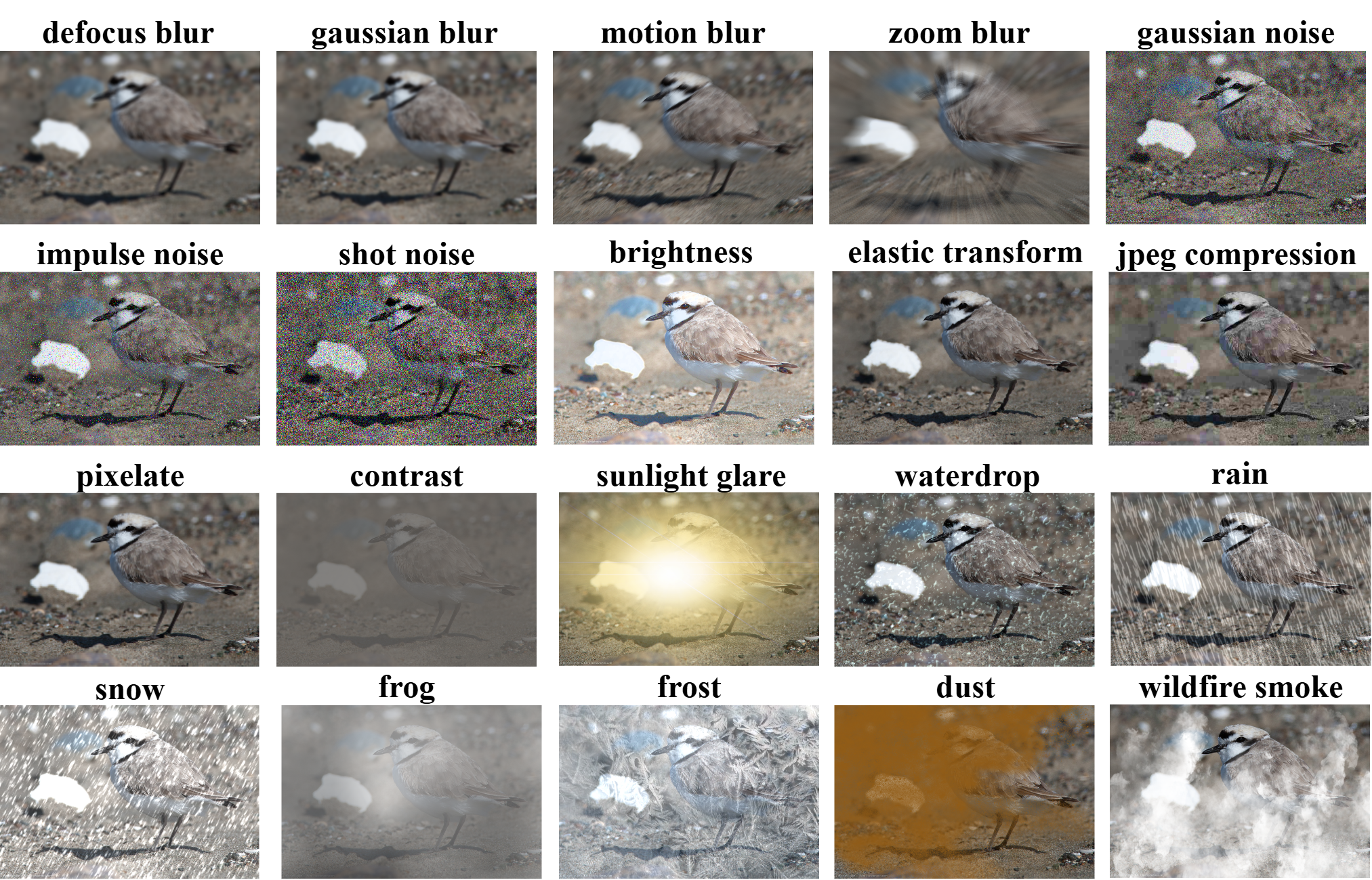}
\captionof{figure}{Visual examples of the 20 image degradations considered in this work. Different degradations affect semantic entities, local structure, and global visual content in different ways.}
\label{fig:20_degradations}
\end{strip}

This section provides supplementary results corresponding to the three constraints of a semantic similarity metric discussed in the main paper, namely, robustness to non-semantic perturbations, sensitivity to entity-level semantic shifts, and sensitivity to relation-level semantic shifts. Specifically, the complete results on the remaining 16 degradation types are used to examine robustness to non-semantic perturbations, the case study on entity changes evaluates sensitivity to entity-level semantic shifts, and the case study on relation changes further tests sensitivity to relation-level semantic shifts. Together, these results provide a more comprehensive validation of the desired properties of T3S.

\subsection{Complete Results on the Remaining 16 Degradation Types}

\begin{figure*}[t] 
\centering 
\includegraphics[width=\textwidth, keepaspectratio]{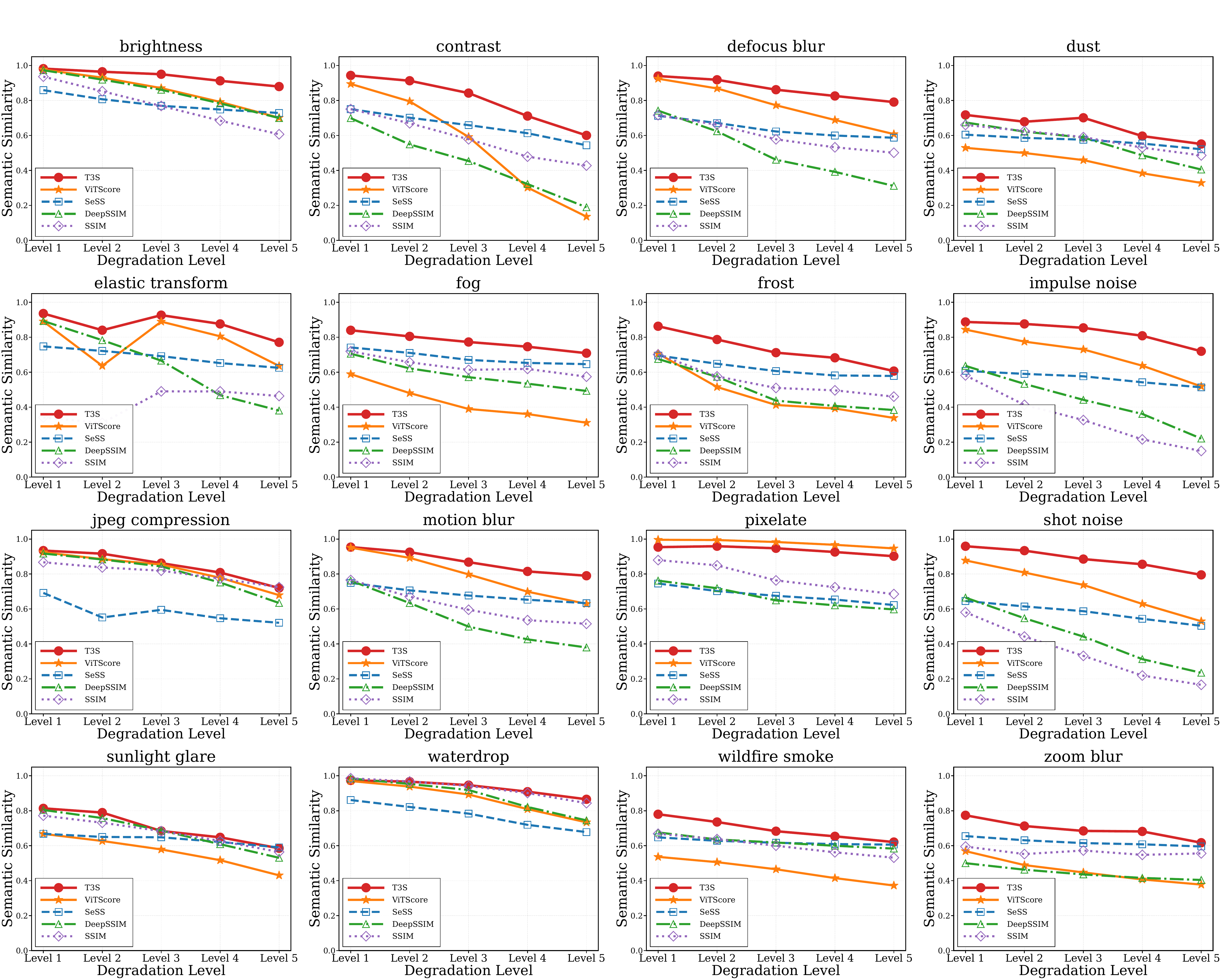}
\caption{Performance curves of different metrics on the remaining 16 degradation types in SPA-Data over five severity levels. Each subfigure shows how the score changes as degradation severity increases.} 
\label{fig:spa_4x4_16curves} 
\end{figure*}

\begin{figure*}[t]
\centering
\includegraphics[width=\textwidth, keepaspectratio]{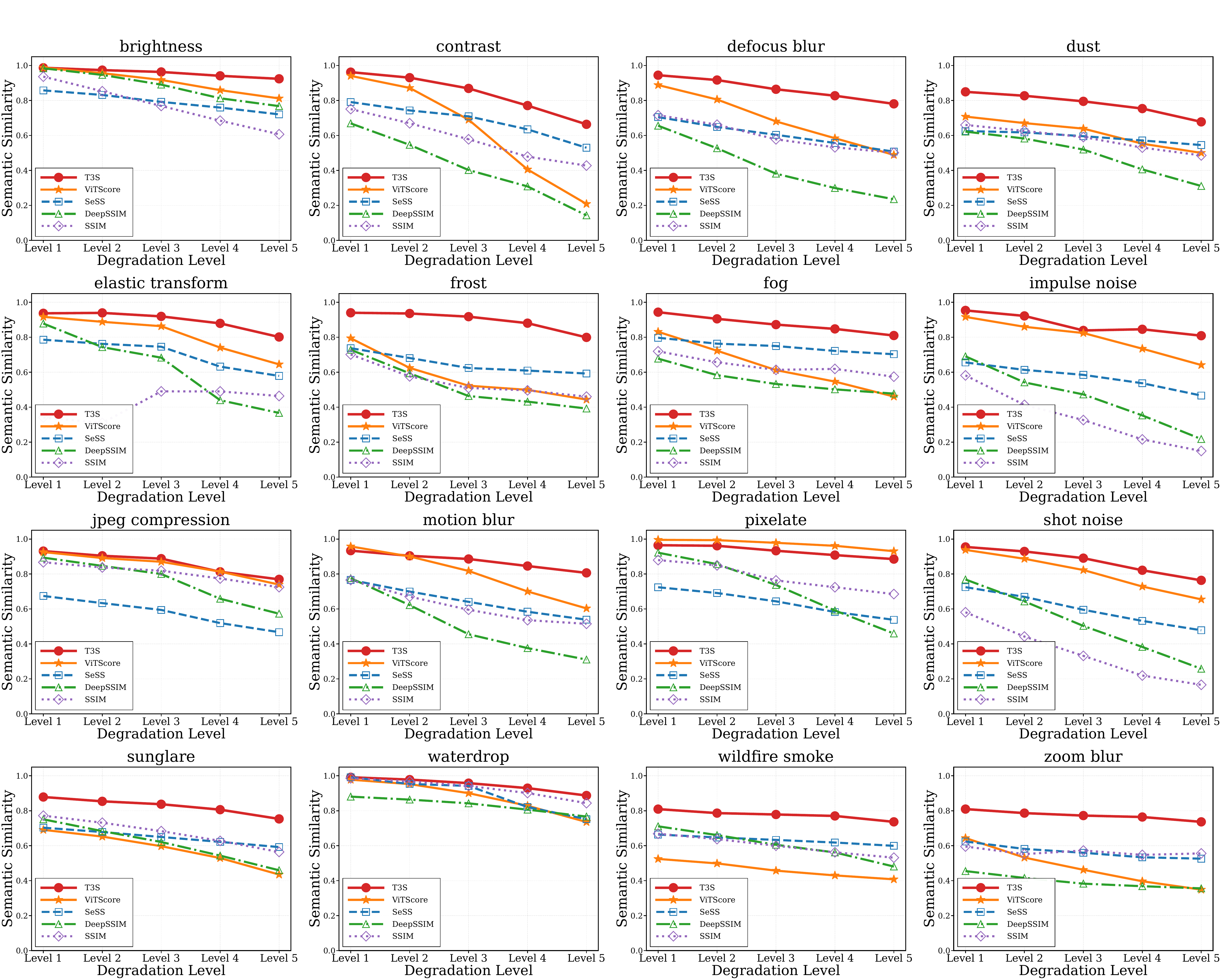}
\caption{Performance curves of different metrics on the remaining 16 degradation types in COCO over five severity levels. Each subfigure shows how the score changes as degradation severity increases.}
\label{fig:coco_4x4_16curves}
\end{figure*}

Figure~\ref{fig:20_degradations} shows 20 visual examples of degraded images, covering diverse low-level distortions with different visual patterns and perceptual effects. For brevity, the main paper reports results on four representative degradation types. Here, we further include the remaining 16 degradation types in our benchmark, thereby completing the evaluation over all 20 degradations considered in this work. This supplementary analysis allows us to examine whether the observations in the main text remain valid across a broader range of low-level distortions.

As shown in Figure~\ref{fig:spa_4x4_16curves}, the $4\times4$ grid of curves for the remaining 16 degradations in SPA-Data follows a largely consistent trend. In most cases, T3S maintains higher scores than the compared baselines and decreases more smoothly as degradation severity increases. From the perspective of semantic robustness, this behavior is desirable: although image quality is progressively reduced, the underlying semantic content is often still preserved, especially at low and medium severity levels. Under such conditions, a reliable semantic similarity index should not treat visual deterioration as a genuine semantic change. The behavior of T3S suggests that it is less likely to confuse non-semantic distortions with true semantic shifts.

Compared with COCO, SPA-Data usually contains fewer salient objects and simpler scenes. Its semantic content is therefore less dense and less structurally complex. In this setting, methods that rely mainly on global features can still maintain high scores as long as the overall layout and coarse scene content remain largely unchanged. This also explains why the performance gap among metrics varies across degradation types in SPA-Data.

This pattern is particularly clear for pixelate. In SPA-Data, increasing pixelation mainly removes local details and blurs object boundaries, but it usually does not alter the overall scene structure or the main object categories. As a result, the semantic difference across severity levels remains limited. Under this distortion, ViTScore stays slightly above T3S, suggesting that global-feature-based metrics are less affected when the overall content is preserved. At the same time, T3S still remains high and decreases gradually with severity, indicating that it can recognize that the image semantics are largely unchanged while still reflecting the progressive loss of local visual quality.

A similar phenomenon can be observed for waterdrop and JPEG compression. These distortions mainly affect local textures, edges, and visual quality, but they do not immediately alter the main scene content. Since SPA-Data contains fewer objects and simpler scenes, metrics such as SSIM and ViTScore can still maintain relatively high scores as long as the main subjects remain recognizable. Therefore, the gap between T3S and these baselines becomes smaller under such distortions. Nevertheless, the relatively stable behavior of T3S supports the same conclusion: it does not overestimate semantic differences when the semantics are in fact preserved.

In contrast, the advantage of T3S is more evident for contrast, shot noise, impulse noise, fog, frost, and defocus blur. These distortions more directly affect the visibility of the main objects and the integrity of their boundaries, but they do not necessarily change the underlying semantics, especially at lower severity levels. For a dataset such as SPA-Data, where image semantics often depend on a small number of key objects, a index that drops too quickly is more likely to misinterpret reduced visibility as semantic inconsistency. By contrast, T3S remains relatively stable on these distortions, which better reflects its robustness to non-semantic perturbations.

As shown in Figure~\ref{fig:coco_4x4_16curves}, the curves on COCO reveal a clear overall trend across these 16 degradations. In most cases, T3S remains above the other methods, and its scores decrease more smoothly as degradation severity increases. Across all 80 score points, T3S ranks first in 71 cases, indicating consistently stronger semantic robustness on most degradations in COCO.

Compared with SPA-Data, COCO usually contains more objects and more complex scenes, making its semantic content richer and more explicit. In this setting, image semantics depend not only on the overall scene, but also on whether multiple local objects remain identifiable and whether their structural relations are preserved. As a result, indices that rely mainly on global information are more likely to overreact to appearance degradation or local visibility loss, even when the core semantics remain unchanged. Under this setting, the advantage of T3S becomes more evident.

This trend is especially clear for dust, fog, frost, wildfire smoke, and sunglare. These degradations obscure the image or reduce the visibility of important regions, but in many cases they do not immediately alter the core semantic content of the scene. On a dataset such as COCO, which contains more objects and more complex scene structure, a index that is overly influenced by visibility degradation may incorrectly treat such quality loss as semantic inconsistency. In contrast, T3S maintains higher scores on these distortions, indicating that it better recognizes that the image semantics are still preserved despite the reduced clarity.

\begin{figure*}[t]
\centering
\includegraphics[width=\textwidth, keepaspectratio]{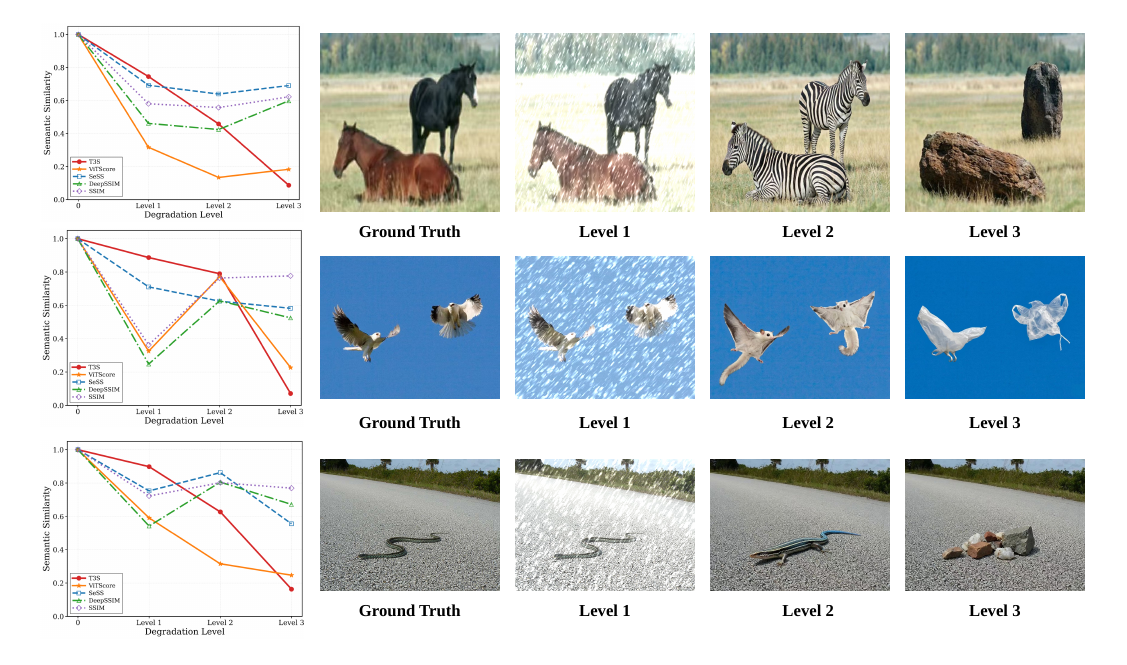}
\caption{Case study on entity-level semantic changes. Each group contains three comparison levels: Level 1 applies strong snow degradation to the original image, Level 2 replaces key entities with semantically related ones, and Level 3 replaces key entities with semantically unrelated ones.}
\label{fig:support_class}
\end{figure*}

A similar pattern can be observed for defocus blur, motion blur, zoom blur, shot noise, and impulse noise. These degradations gradually damage textures, edges, and local structures, but at low to medium severity, the main objects and their relations are often still recognizable. On a semantically richer dataset such as COCO, many baselines drop rapidly because they are overly sensitive to local appearance changes. By contrast, T3S decreases more slowly, suggesting that it can better distinguish a loss of visual quality from a true semantic change.

There are still a few exceptions. For example, on pixelate, ViTScore remains slightly higher than T3S at all five severity levels. This is because pixelation mainly affects local textures and boundaries, while the overall scene structure and the main object categories usually remain unchanged. As a result, methods that focus more strongly on global information can more easily retain high scores. Even so, T3S still stays at a relatively high level and decreases gradually as degradation becomes stronger, which remains consistent with the hypothesis that the semantics are largely preserved under this distortion.

Taken together, the results on COCO and SPA-Data reveal a consistent overall pattern: for most degradation types, T3S maintains relatively high scores and decreases more smoothly as degradation severity increases. This suggests that T3S is better able to distinguish visual quality deterioration from genuine semantic change across both datasets.

The main difference between the two datasets lies in their semantic complexity. SPA-Data usually contains fewer objects and simpler scenes, so global-feature-based methods such as ViTScore can more easily maintain high scores as long as the overall image remains visually similar. This is why the gap among methods is smaller for degradations such as pixelate, waterdrop, and JPEG compression. By contrast, COCO contains more objects and more complex scenes, so local object quality and structural relations play a more important role in semantic consistency. Under this setting, the robustness advantage of T3S becomes more pronounced.

A shared pattern across both datasets is that pixelate serves as a special case. On both COCO and SPA-Data, ViTScore remains very high under this distortion and is usually slightly higher than T3S. This suggests that pixelation mainly affects local details rather than the overall content or layout of the image. In contrast, T3S still remains high and decreases gradually with severity, which is more consistent with the view that image semantics are largely preserved while visual quality is progressively degraded.

\subsection{Case Study on Entity Changes}
\begin{figure*}[t]
\centering
\includegraphics[width=\textwidth, keepaspectratio]{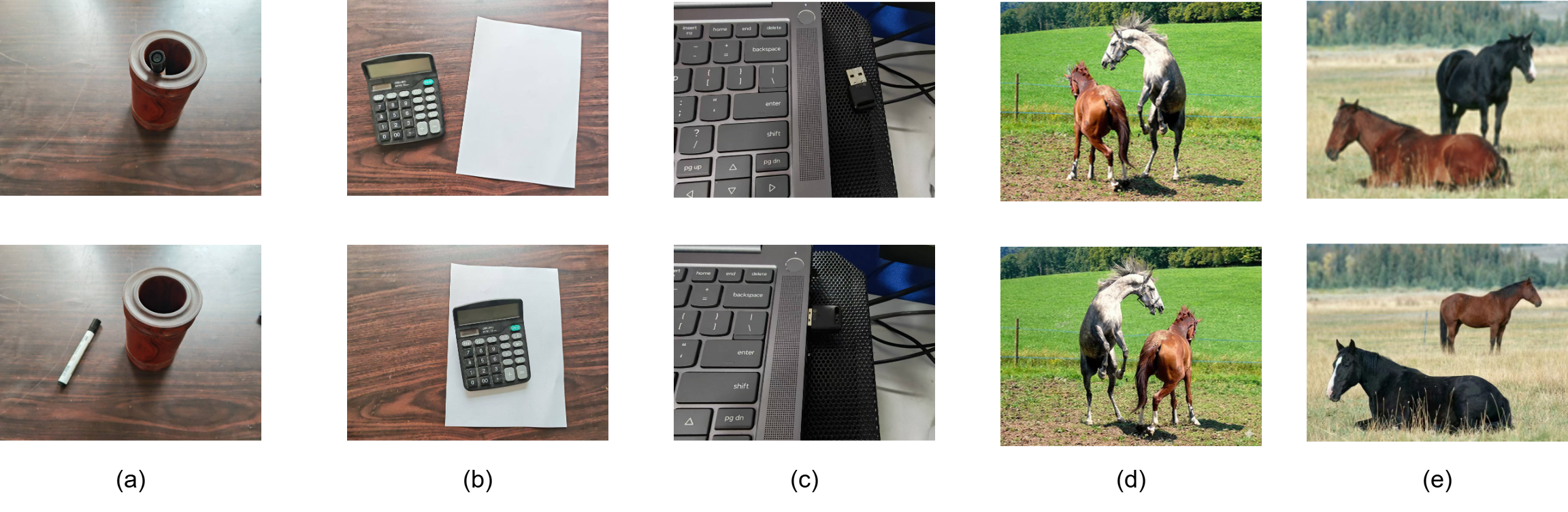}
\caption{Examples of image pairs with relation-level semantic changes. In each pair, the main semantic entities are largely preserved, while the relations among them are changed. The first three pairs, (a)--(c), are real photographs, and the last two pairs, (d) and (e), are generated examples. The corresponding T3S scores are: (a) 0.7641, (b) 0.8062, (c) 0.8231, (d) 0.8316, and (e) 0.8293.}
\label{fig:relation}
\end{figure*}

To further examine sensitivity to entity-level semantic changes, we construct a small set of image pairs in which one or more key entity categories are modified using a generative editing model. Figure~\ref{fig:support_class} presents these examples for qualitative analysis only, since the number of controllable generated cases is limited and does not support a large-scale benchmark.

We design a three-level comparison setting. Level 1 applies a strong snow degradation to the original image. Although the image quality is noticeably reduced, the underlying semantic content remains unchanged. Level 2 replaces key entities with semantically similar ones, such as replacing a horse with a zebra or a snake with a lizard. This introduces a moderate semantic shift while preserving part of the original semantic meaning. Level 3 replaces key entities with semantically unrelated ones, such as replacing a bird with a plastic bag or replacing a horse and a snake with stones. In this case, the semantic content is fundamentally changed.

An appropriate semantic similarity index should therefore produce the following ordering: the highest score for Level 1, a lower but still moderate score for Level 2, and the lowest score for Level 3. Among the compared methods, only T3S consistently preserves this ordering across all three groups of examples. Specifically, T3S follows a clear monotonic pattern, i.e., Level 1 $>$ Level 2 $>$ Level 3, in every group. This indicates that T3S can correctly recognize that heavy snow mainly affects visual quality rather than semantics, assign an intermediate similarity to semantically related substitutions, and sharply reduce the score when the replaced entities are semantically unrelated.

This trend is also reflected in the average results across the three groups. The average T3S scores are 0.8436, 0.6251, and 0.1073 for Levels 1, 2, and 3, respectively, showing a clear progressive decline with increasing semantic discrepancy. In contrast, several baselines do not preserve such semantic ordering. For example, the average score of DeepSSIM increases from 0.4171 at Level 1 to 0.6190 at Level 2, and the average score of SSIM increases from 0.5553 to 0.7079, suggesting that these fidelity-oriented indices are driven more by low-level appearance similarity than by semantic consistency. ViTScore yields similar average scores for Levels 1 and 2 (0.4107 vs. 0.4075), indicating limited ability to distinguish non-semantic degradation from semantically related entity substitution. SeSS also remains relatively high across all three levels (0.7188, 0.7091, and 0.6097), but its decline is much smaller and less discriminative than that of T3S.

A more important observation is that T3S not only preserves the correct ranking, but also assigns substantially different penalties to different semantic shifts. The average score drop of T3S from Level 1 to Level 2 is 0.2185, whereas the drop from Level 2 to Level 3 reaches 0.5178. This means that T3S imposes a much stronger penalty when the semantic replacement changes from related to unrelated, which is consistent with human intuition about semantic similarity. By contrast, SeSS decreases by only 0.0097 from Level 1 to Level 2 and by 0.0994 from Level 2 to Level 3 on average, showing much weaker sensitivity to the degree of semantic mismatch.

Overall, these results demonstrate that T3S better matches the desired behavior of a semantic similarity index. It remains high when image semantics are preserved despite severe visual degradation, decreases moderately when entities are replaced by semantically related ones, and drops sharply when the replacements are semantically unrelated. This suggests that T3S is more aligned with graded semantic change than with low-level appearance similarity alone.

\subsection{Case Study on Relation Changes}

Figure~\ref{fig:relation} presents five image pairs in which the main semantic entities are largely preserved, while the relations among them are changed. The first three pairs are real photographs, whereas the last two are generated examples constructed to emphasize relation-level semantic changes. In these cases, the T3S scores are still reduced, indicating that the proposed metric can perceive relation-level discrepancies even when the entity categories remain unchanged. At the same time, the scores remain relatively high, typically around 0.8, suggesting that such changes are captured only weakly and are penalized less strongly than entity-level substitutions.

A main reason is the limited ability of Qwen2.5-VL-3B to capture fine-grained relational differences. When two images are jointly input, if the object categories remain unchanged, the model tends to produce the same or very similar relations for both images. Consequently, the relation branch receives only weak discrepancy signals, and the final score is influenced more by the global semantic feature than by explicit relation inconsistency.

For example, in Figure~\ref{fig:relation}(d), the two images contain the same object categories, namely two horses, but their relative positions are clearly different. However, Qwen may still output the same relation, such as \textit{horse next to horse}, for both images. As a result, the actual relation change is only weakly reflected in the relation representation, and the final score remains relatively high. Similar behavior can also be observed in the other examples: although the spatial arrangement or interaction pattern changes, the extracted relation texts remain highly similar because the entity set itself is largely unchanged.

Importantly, a relatively high similarity score is not entirely unreasonable in these cases, since the main entities and the overall scene content are still largely preserved. Rather, these examples show that relation-level changes are subtler than entity-level substitutions: they can be recognized by T3S, but the current relation extraction module does not yet impose a sufficiently strong penalty. This further suggests that, although relation modeling is necessary for semantic similarity assessment, accurately capturing subtle relation-level changes remains an open challenge for future improvement.

\section{Discussion on Limitations and Future Work}

Despite the promising results, the current framework still has several limitations. First, the Qwen-based semantic extraction module is relatively large and introduces substantial computational overhead, which makes inference slow and limits the practical efficiency of the overall index. This issue becomes more pronounced when a large number of image pairs must be evaluated. Second, the performance of the current Qwen model is still insufficient for fine-grained semantic understanding, especially in capturing subtle relation-level changes and maintaining consistent predictions across image pairs. As a result, some semantic discrepancies may be underestimated, particularly when the entity categories remain unchanged but their spatial or interactive relations differ.

In future work, we plan to improve the framework from both efficiency and capability perspectives. On the one hand, we will explore lighter and faster semantic extraction models, or distill the current large model into a compact version, to reduce inference cost and improve scalability. On the other hand, we will investigate stronger relation-aware semantic modeling methods to better capture subtle entity and relation changes. A more accurate and efficient semantic extraction module would further enhance the reliability of T3S and make it more practical for large-scale evaluation and potential training-time applications.

\end{document}